\documentclass[letterpaper, 10 pt, conference]{ieeeconf}

\IEEEoverridecommandlockouts
\usepackage{graphics} 
\usepackage{epsfig} 
\usepackage{mathptmx} 
\usepackage{times} 
\usepackage{amsmath} 
\usepackage{amssymb}  
\usepackage{booktabs}
\usepackage{siunitx}
\usepackage{caption}
\usepackage{multirow}
\usepackage{url}
\usepackage{svg}
\usepackage{balance}

\newcommand{\up}{$\scriptstyle\uparrow$}
\newcommand{\down}{$\scriptstyle\downarrow$}

\title{\LARGE \bf
SpectralWaste Dataset: Multimodal Data for Waste Sorting Automation
}

\author{Sara Casao$^{1,*}$\qquad 
Fernando Peña$^{1,*}$\thanks{${}^*$Authors contributed equally to this work.}\qquad 
Alberto Sabater$^{1}$\qquad\\
Rosa Castillón$^{2}$\quad 
Darío Suárez$^{1}$\qquad
Eduardo Montijano$^{1}$\qquad
Ana C. Murillo$^{1}$\qquad
\thanks{$^{1}$Universidad de Zaragoza}%
\thanks{$^{2}$ATRIA Innovation}%
\thanks{This work was supported by DGA project T45\_23R and by MCIN/AEI/ERDF/European Union NextGenerationEU/PRTR project PID2021-125514NB-I00.}
}

\begin{document}

\maketitle
\thispagestyle{empty}
\pagestyle{empty}

\begin{abstract}
The increase in non-biodegradable waste is a worldwide concern. Recycling facilities play a crucial role, but their automation is hindered by the complex characteristics of waste recycling lines like clutter or object deformation.
In addition, the lack of publicly available labeled data for these environments makes developing robust perception systems challenging.
Our work explores the benefits of multimodal perception for object segmentation in real waste management scenarios. 
First, we present SpectralWaste, the first dataset collected from an operational plastic waste sorting facility that provides synchronized hyperspectral and conventional RGB images. This dataset contains labels for several categories of objects that commonly appear in sorting plants and need to be detected and separated from the main trash flow for several reasons, such as security in the management line or reuse. 
Additionally, we propose a pipeline employing different object segmentation architectures and evaluate the alternatives on our dataset, conducting an extensive analysis for both multimodal and unimodal alternatives. Our evaluation pays special attention to efficiency and suitability for real-time processing and demonstrates how HSI can bring a boost to RGB-only perception in these realistic industrial settings without much computational overhead. 
\end{abstract}

\section{Introduction}
The global issue of waste production intensifies as societies grow and consumption rises. The sheer volume of waste generated, particularly non-biodegradable waste such as plastics, has reached concerning proportions~\cite{sukno2022hand}.
Recycling and reuse are key strategies to lessen the environmental burden of waste. Hence, automating waste management facilities not only increases the volume of properly processed waste but also safeguards worker health and comfort. 
To achieve automated manipulation of relevant elements in these real industrial environments, the first step is to improve and adapt existing perception systems to this domain. 

Despite the great advances in automated visual recognition tasks in recent years, real industrial settings often present recurring problems that hinder real-world applicability, such as the lack of annotated data to achieve precise domain adaptation or high computational requirements~\cite{linder2021lack}. The most common method to identify and localize elements of interest in automated tasks is through the segmentation of RGB images~\cite{lee2022maskgrasp, lu2023vl}. However, accurate detection based only on visual features is extremely challenging in waste management scenarios with severe clutter, high materials diversity, deformable or broken objects, and translucent elements (see sample images in Figure~\ref{fig:mosaicdata}). 

To overcome these issues, the use of more complex sensing modalities like hyperspectral imaging (HSI), commonly used for raw material classification~\cite{henriksen2022plastic}, can provide insights beyond the visual appearance of objects.
While conventional RGB cameras capture the visible spectrum, hyperspectral cameras are able to acquire light across a wide range of wavelengths.
Thus, leveraging the combined information from both modalities improves the performance of complex perception tasks such as segmentation of buildings~\cite{habili2022build_multimodal}, terrain classification~\cite{kodgule2019terrain} or object tracking~\cite{liu2022siamhyper}. Unfortunately, the advantages derived from using hyperspectral information along with RGB images remain unexplored in object recognition for automatic waste sorting, where the task is mostly approached using RGB information~\cite{bashkirova2022zerowaste}.

\begin{figure*}[!bt]
    \centering
    \includegraphics{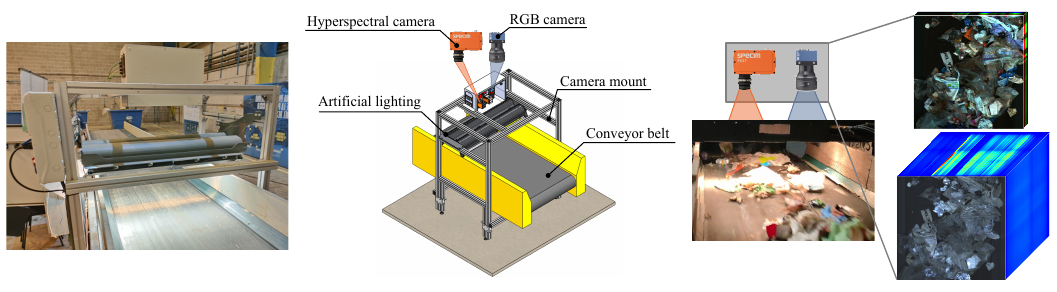}
    \caption{The multimodal setup in the waste sorting facility contains two synchronized line-scan cameras that gather RGB and hyperspectral data. \textit{Left:} Real prototype installed in the facility. \textit{Middle:} Diagram of the setup for data capture. Both hyperspectral and RGB cameras are housed in an industrial enclosure and completed with artificial lighting to ensure correct image capture. \textit{Right:} Example of a scene as captured by the RGB camera and by the hyperspectral camera (with a false-color for visualization purposes).}
    \label{fig:prototype2}
\end{figure*}

This work demonstrates the benefits of using multimodal segmentation approaches in waste management and contributes to mitigating two key challenges that currently hinder their adoption, the scarcity of public multimodal waste datasets and the high computational demands associated with HSI data. 
Specifically, our main contributions are twofold: 
(1) We introduce SpectralWaste\footnote{Dataset website: \url{https://sites.google.com/unizar.es/spectralwaste}. Models and code will be released upon acceptance.}, the first multimodal dataset obtained from an operational waste sorting facility, featuring in-the-wild industrial data from both hyperspectral and RGB cameras (Figure~\ref{fig:prototype2}). This dataset addresses the identification of critical objects that frequently appear in real trash flows and impact sorting efficiency by either clogging machinery if not removed or holding value if recovered.
(2) We present a comprehensive object segmentation analysis that underscores the boost in performance when combining both modalities and, for the first time, explores the suitability of using HSI for object segmentation in waste sorting scenarios. The proposed pipeline places particular emphasis on employing efficient architectures. Furthermore, to ensure consistency in annotated masks between modalities and reduce the labeling effort required, we propose a novel label transfer algorithm that automatically adapts RGB-annotated masks to HSI without any calibration needed.

\section{Related Work}
This section summarizes existing works regarding waste datasets for object identification, and image segmentation methods using both hyperspectral imaging exclusively and multimodal information.

\subsection{Waste Object Identification Datasets}
Numerous datasets spanning diverse domains have focused on collecting data for waste identification. Stanford TrashNet~\cite{yang2016simpledataset} consists of classifying single objects on an empty white background. Aiming to tackle the localization problem in addition to classification, Trash Annotations in Context (TACO)~\cite{proenca2020taco} presents a dataset in open and real environments, such as streets, lakes, or beaches where world litter is shown. Following the idea of reducing waste in natural environments, Floating Waste (FloW)~\cite{cheng2021flow} is dedicated to the efficient cleaning of inland water areas with autonomous boats. Due to the demands of this task, a multimodal sub-dataset, FloW-RI, is included, providing millimeter wave radar data synchronized with the images. Similarly, other works aim to tackle challenging tasks by providing complementary information alongside RGB images. For instance, \cite{kim2023transpose} introduces a multimodal dataset comprising RGB-D, thermal infrared and object poses to address the issue of transparent object identification. 

Closer to our work is the ZeroWaste dataset~\cite{bashkirova2022zerowaste} and its extension ZeroWaste-v2, proposed in the VisDA2022 challenge~\cite{bashkirova2023visda}. In these papers, they provide a dataset comprised of conventional RGB images for industrial waste object segmentation that have been collected from a real sorting plant. 
In contrast, our dataset includes hyperspectral data synchronized with the RGB images. Thus, this spectral information combined with data from the visible spectrum holds significant potential for enhancing object identification tasks within the recycling processes. 

\subsection{Identification with Hyperspectral Data}
A wide variety of techniques have been explored to leverage hyperspectral sensors for raw material identification, including spectral angle mapper~\cite{kruse1993spectral}, multi-layer perceptrons~\cite{hanson2023slurp} and CNNs~\cite{seidlitz2022robust}. However, the limited amount of data in the existing HSI datasets poses a challenge for training data-based models. Common methods involve pixel-wise training and testing on a reduced set of images~\cite{wendel2016self}, leading to information leakage and suboptimal generalization capabilities~\cite{nalepa2019validating}.

Final applications with HSI typically focus on fields where the information captured by RGB cameras lacks sufficient detail, e.g., environmental monitoring, agriculture, medical imaging, or remote sensing~\cite{grewal2023hyperspectral}. More specifically, the use of hyperspectral sensors for automatic plastic sorting in recycling facilities is a widespread technique. For example, the structure of plastic material is analyzed by sparse pixels~\cite{shiddiq2023plastic, henriksen2022plastic}, or hyperspectral imaging is used to densely label images based on per-pixels classifications~\cite{karaca2013automatic}. Unlike these works focused on raw material identification, we study the use of HSI for object segmentation, which can potentially overcome the limitations of RGB data in waste sorting tasks by leveraging spectral properties for material differentiation.

Regarding multimodal sensors for segmentation tasks, multiple works have addressed this problem through sensor fusion~\cite{chen2020geomorphological, valada2017adapnet}. In particular, the combination of HSI with RGB information has been used in multiple fields. For instance, combining HSI with RGB among other modalities is exploited in environmental monitoring with UAV's~\cite{qin2022trees_multimodal} and autonomous terrain classification~\cite{kodgule2019terrain}. Moreover, perception tasks combine both modalities for different tasks like classification of building materials~\cite{habili2022build_multimodal} or rise seeds inspection~\cite{fabiyi2020seeds_multimodal}. However, the exploration of multimodal segmentation in real-world industrial waste sorting scenarios remains an open area of research. 

\section{SpectralWaste Dataset}
\label{sec:dataset}
This section describes the novel multimodal SpectralWaste dataset, explaining the data acquisition followed and the annotation process performed. 

\begin{figure*}
    \centering
    \includegraphics[width=\textwidth]{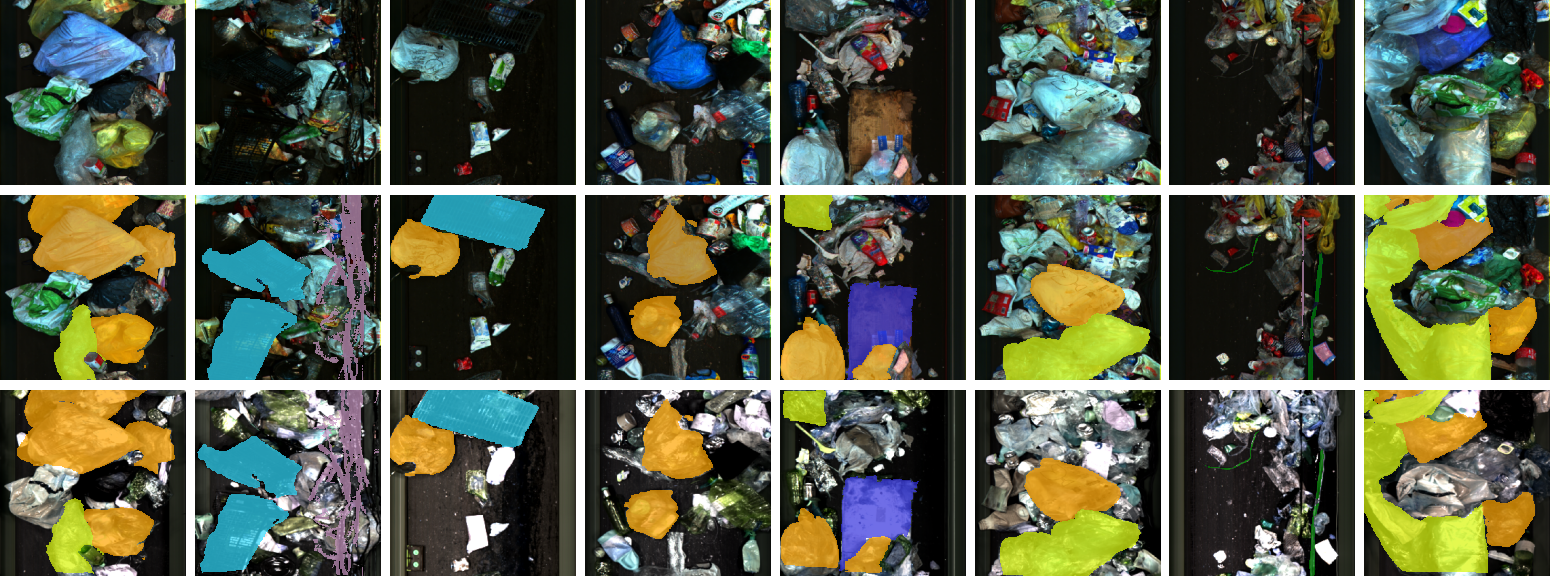}
    \includegraphics{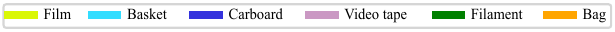}
    \caption{Examples of images included in the dataset. The first row shows images captured by the RGB camera, the second row shows the ground-truth RGB annotations and the third row shows the same scenes captured by the hyperspectral camera, annotated with labels obtained using the proposed label-transfer algorithm. For visualization purposes, we manually select three out of all the hyperspectral bands.}
    \label{fig:mosaicdata}
\end{figure*}

\subsection{Data Acquisition} 
The dataset was collected in a real waste sorting industry specialized in plastics, cartons and cans, with a true-to-life prototype of the conveyor belt installed on the waste separation line (see Figure~\ref{fig:prototype2}). This prototype closely mimics the real installation, ensuring that the captured waste streams accurately mirror those arriving at the facility for separation.

The setup involved two synchronized cameras for multimodal data capture: a line-scan RGB camera (Teledyne~DALSA~Linea) and line-scan hyperspectral sensor (Specim~FX17) that captures 224 contiguous spectral bands in a range from~900~to~1700~nm.
Both cameras were housed in an industrial enclosure situated at a height of 1.7m. This installation was supplemented with a set of LED illuminators and infrared halogen lighting to ensure a suitable image capture in the spectral domain.
RGB images were stored with a resolution of $1200\times1184$ pixels and 8-bit color depth, while HSI images were stored as data cubes of size $600\times640\times224$ with 16-bit precision. 

\subsection{Data Annotation}
\label{sec:ann}
The classes chosen for annotation in the presented dataset were selected according to the requirements of the facility. Labeled objects represent elements that commonly cause operational problems in recycling lines, impacting the efficiency of the sorting process. Among these problems, machinery jams pose a significant issue, causing a complete stoppage of the waste separation until the obstructing object is removed. Thus, the selected objects for automatic identification include \textit{film} and \textit{basket}, large objects that can clog the conveyor belts as they are not easily breakable; \textit{video tape} and \textit{filament}, representing long objects prone to entangle in waste separation zones and requiring manual intervention; \textit{trash bag}, which encompasses closed bags containing waste that need to be mechanically opened for further processing; and \textit{cardboard}, paper objects received at the facility whose recovery adds value sent to another recycling process. 

To streamline the time-consuming process of labeling and to facilitate this tedious task, we developed an interactive segmentation tool\footnote{Code will be released upon acceptance.} leveraging the point-prompt feature from Segment Anything Model (SAM)~\cite{kirillov2023sam}. In essence, the user can select points on the image belonging to an object, and with each selection, a new mask is generated and displayed over the image for further refinements or saving. In this work, we used our tool to manually generate the ground truth masks of the defined objects in the RGB image set. 

\subsection{Dataset Content} 
The result of the entire data acquisition and annotation process is encompassed in the SpectralWaste dataset. The dataset provides annotations for six object classes: \textit{film, basket, video tape, filaments, trash bags}, and \textit{cardboard}, totaling 2059 annotated instances across a set of 852 non-overlapping images. Table~\ref{tab:dataset} presents the overview of the annotations, while Figure~\ref{fig:mosaicdata} and~\ref{fig:mosaicdata2} illustrate sample images from the different classes. 

In addition to the labeled set, SpectralWaste contains 6803 unlabeled multimodal images (RGB- HSI). We believe that releasing these unlabeled images is valuable for the community, enabling the researchers to explore the advantage of hyperspectral or multimodal object identification in a real industrial waste facility through further study of different techniques. These techniques may include refining labeled segmentation using semi-supervised or self-supervised methods, as well as exploring unsupervised segmentation approaches.  

\begin{table}[!ht]
\caption{Summary of the instances annotated in SpectralWaste.}
\centering
\footnotesize
\begin{tabular}{ccccccc}
\toprule
& \multicolumn{6}{c}{\textbf{Instances per class}} \\
\cmidrule(r){2-7}
\textbf{Total} & \emph{Film} & \emph{Basket} & \emph{Card.}  & \emph{Tape} & \emph{Filam.} & \emph{Bag} \\
\midrule
2059 & 339 & 300 & 68 & 287 & 111 & 954 \\
\bottomrule
\end{tabular}
\label{tab:dataset}
\end{table}

\section{Waste Segmentation}
\label{sec:wsAlg}
\begin{figure}[!t]
    \centering
    \includegraphics[width=\linewidth]{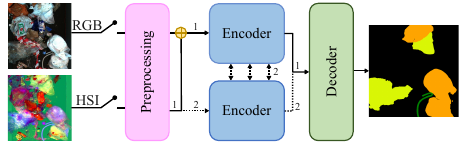}
    \caption{Pipeline of the segmentation process. Path 1 represents the data flow for the initial architectures designed for unimodal RGB segmentation (MiniNet-v2 and SegFormer). The dashed path 2 represents the HSI data flow in the multimodal CMX architecture.}
    \label{fig:baselines_diagram}
\end{figure}

This section describes the proposed pipeline for waste object segmentation using RGB and HSI images. Figure~\ref{fig:baselines_diagram} summarizes the key steps of the process where we consider different configurations that can take one or both modalities.
A detailed description of the steps involved in the baselines and the adapted architectures is provided in the following.

\subsection{Data Preprocessing}
\label{sec:preprocess}

\paragraph*{General preprocessing.}
The pipeline includes a series of common preprocessing steps. We first crop the mismatched areas captured by each sensor to align the space shown in each image, resize the RGB and HSI images to $256\times256$ pixels and scale them to 32-bit floating-point values in the range $[0,1]$. 

\paragraph*{HSI channel reduction.}
Hyperspectral imaging imposes significant computational demands due to the large amount of information contained in each pixel. Aiming to explore efficient solutions within our pipeline, we implement dimensionality reduction across the spectral channels using principal component analysis (PCA) to select the three main components (HYPER3). This PCA analysis is conducted on the pixel values from the training set of the dataset and then applied to the validation and testing sets to obtain their reduced version. The reduction process maintains 99.7\% of the explained variance over the training set. By reducing the HSI channels to three components, we not only obtain a compressed representation of the information but also provide a balanced input when combining HSI with RGB images in the multimodal configurations

\subsection{Segmentation Architectures}
Considering the application addressed in this work of segmenting objects in an industrial setting, we compare three different architectures within our pipeline paying special attention to alternatives suitable for real-time inference. 

The first architecture is \textit{MiniNet-v2}~\cite{alonso2020mininet}, a lightweight convolutional neural network designed for segmentation tasks. This model uses multi-dilation depthwise separable convolutions and two convolutional branches instead of skip connections to achieve a favorable trade-off between accuracy and computation. 
The second architecture is \textit{SegFormer}~\cite{xie2021segformer}, a well-known transformer-based segmentation network. Specifically, we opted for the version with the smallest encoder (SegFormer-B0) as it is reported to be suitable for real-time environments and closer to MiniNet-v2 in terms of computational requirements and number of parameters. Since both MiniNet-v2 and SegFormer were originally designed for processing only RGB images, we adapt them for multimodal segmentation by early-fusing both modalities. This early fusion involves concatenating the RGB and HSI images across the channel dimension before feeding them into the network (see path 1 in Figure~\ref{fig:baselines_diagram}). Finally, we evaluate \textit{CMX}~\cite{zhang2023cmx}, a hybrid-fusion transformer based on SegFormer. This network receives RGB and HSI data separately, processing them with two encoders that share information throughout the network (see path 2 in Figure~\ref{fig:baselines_diagram}). This model also integrates feature rectification techniques to mitigate the effects of noisy measurements from different modalities, which is crucial in our case.
It is noteworthy that, while CMX has been previously assessed with multiple modalities (depth, polarization, event and LiDAR), including multispectral ones (thermal and infrared)~\cite{wang2023tirdet, yan2023cross, chen2023igt}, it has not been evaluated on HSI data before.

In summary, our pipeline offers flexibility in handling different data types. MiniNet-v2 and SegFormer can be used for processing individual RGB or HSI images (unimodal inputs). Additionally, all three architectures, i.e., MiniNet-v2, SegFormer, and CMX, can be employed for multimodal segmentation combining RGB and HSI data.

\subsection{Label Transfer}
To explore the unimodal HSI configurations of our pipeline, we address the challenge of data misalignment, i.e., the inaccurate alignment of manually RGB-annotated masks to the corresponding objects in HSI. This effect is due to the physical distance between the cameras and the heterogeneous internal settings of each sensor, which causes the perspective captured by each camera to differ.
Since the employed cameras are non-conventional (line-scan cameras) and we only have two views with no additional spatial information such as depth, the well-established calibration methods for pinhole cameras are not applicable~\cite{behmann2015calibration}. Therefore, to ensure consistency in annotated masks between modalities while improving labeling efficiency, we propose a novel label transfer algorithm.
Our approach automatically adapts annotated segmentation masks from one camera (RGB) to another (HSI), relying exclusively on both images. The objective is to find affine transformations per mask that adapt the existing segmentation to the size and perspective of the corresponding object in the target image.

\begin{figure}[t]
    \centering
    \includegraphics{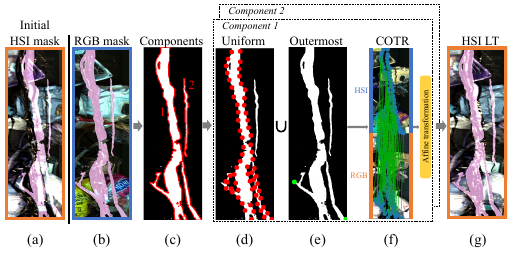}
    \caption{Visualization of the main steps of the proposed label transfer method: (a) initial mask superposed to HSI; (b) original manually annotated mask (in pink) superposed to RGB image; (c) contours of the extracted connected components; (d) uniformly-sampled points (red); (e) outermost points (green); (f) COTR matching for selected points in RGB (top) to the HSI image (bottom) used to compute an affine transformation for the mask; (g) resulting mask after projection into the HSI image.}
    \label{fig:labeltransfer}
\end{figure}

The different steps of the algorithm are visualized in Figure~\ref{fig:labeltransfer}. First, we extract the contours of the segmentation mask and process each connected component independently. For example, in Figure~\ref{fig:labeltransfer}(c) two different local affine transformations are computed, one per component. 
Then, we sample each contour to obtain a sparse representation of the shape, generated by uniformly-sampled points (Figure~\ref{fig:labeltransfer}(d)), always including the outermost points for better representativeness (Figure~\ref{fig:labeltransfer}(e)).  
The next step involves identifying the corresponding points in the target image.
To accomplish this, we leverage the feature matching system COTR~\cite{jiang2021cotr}, which uses a transformer-based model
 to find the point on a target image that corresponds to a query point on a source image (Figure~\ref{fig:labeltransfer}(f)).
Finally, we obtain the affine transformation from both sets of points and apply it to the component.
The final mask is obtained by combining all the transformed components (Figure~\ref{fig:labeltransfer}(g)). Compared to the initial mask in Figure~\ref{fig:labeltransfer}(a), its alignment with the underlying object is significantly improved.

\section{Experiments}
This section presents several experiments to evaluate the waste segmentation pipeline (Section~\ref{sec:wsAlg}) in the SpectralWaste dataset (Section~\ref{sec:dataset}).

\subsection{Experimental Settings}  
\paragraph*{Training configuration}
The training of our pipeline runs with the following configuration based on the recommendations from the original works~\cite{alonso2020mininet,xie2021segformer,wang2023tirdet}: MiniNet-v2 uses the Adam optimizer with \num{1e-3} initial learning rate, \num{1e-4} weight decay and a polynomial schedule with the power set to 0.9. SegFormer models are trained using AdamW with \num{1e-3} initial learning rate and a polynomial scheduler with 0.1 power. The selected CMX architecture is the one based on SegFormer and is trained using AdamW with \num{1e-3} initial learning rate and a polynomial schedule with power 0.9. In all cases, the loss is calculated using the softmax cross entropy function and the models are trained during 200 epochs. We weight the classes in the loss calculation using median pixel-level frequency balancing. Regarding data augmentation, the training set is augmented with random rotations of $\pm30$ degrees and vertical and horizontal flips.

\paragraph*{Metrics}
To evaluate the implemented architectures, preprocessing steps, and fusion methods, the intersection over union (IoU) per class is computed. As the final metric, we report their mean (mIoU).

\begin{table*}[!tb]
\caption{Evaluation of the annotations alignment between modalities of manual alignment (MA) and automatic label transfer (LA). Results are computed with the IoU of the obtained masks with a set of 81 instances manually labeled in a subset of 20 hyperspectral images.}
\footnotesize
\centering
\begin{tabular}{lccccccc}
\toprule
& \multicolumn{6}{c}{\textbf{IoU (\%)} \up} & \\
\cmidrule(r){2-7}
\textbf{Alignment Method} & \textit{Film} & \textit{Basket} & \textit{Cardboard} & \textit{Video tape} & \textit{Filament} & \textit{Trash bag} & \textbf{mIoU (\%) \up} \\
\midrule
Manual Alignment (MA) &
69.0 & 61.1 & 81.2 & 27.4 & 25.1 & 74.0 & 56.3 \\
Label Transfer (LT) &
78.7 & 78.3 & 93.5 & 57.1 & 81.2 & 88.8 & 79.6 \\
\bottomrule
\end{tabular}
\label{tab:manual_vs_transfer}
\end{table*}

\begin{table*}[!tb]
\caption{Hyperspectral image segmentation with different supervision: labels resulting from manual alignment (MA) and the proposed automatic label transfer (LT). Note how the LT labels help to obtain better models, particularly for the very thin objects that were completely missed by the MA labels.}

\centering
\footnotesize
\begin{tabular}{lllcccccc c}
\toprule
& & & \multicolumn{6}{c}{\textbf{IoU (\%)} \up} \\
\cmidrule(r){4-9}
\textbf{Backbone} & \textbf{Modality} & \textbf{Labels} &\textit{Film} & \textit{Basket} & \textit{Cardboard} & \textit{Video tape} & \textit{Filament} & \textit{Trash bag} & \textbf{mIoU (\%) \up} \\
\midrule
MiniNet-v2 & HYPER & MA & 59.2 & 57.2 & 76.2 & 17.2 & 19.2 & 49.2 & 46.3 \\
MiniNet-v2 & HYPER & LT & 61.2 & 61.0 & 78.8 & 28.8 & 30.5 & 56.3 & 52.8 \\
MiniNet-v2 & HYPER3 & MA & 56.8 & 52.8 & 62.9 & 19.2 & 6.9 & 45.9 & 40.7 \\
MiniNet-v2 & HYPER3 & LT & 58.8 & 61.9 & 69.4 & 30.2 & 23.0 & 50.5 & 49.0 \\
\midrule
SegFormer-B0 & HYPER & MA & 61.8 & 59.1 & 85.0 & 18.6 & 26.3 & 52.0 & 50.5 \\
SegFormer-B0 & HYPER & LT & 65.4 & 63.2 & 85.2 & 21.9 & 33.1 & 57.2 & 54.3 \\
SegFormer-B0 & HYPER3 & MA & 56.6 & 55.4 & 84.7 & 14.4 & 27.5 & 46.7 & 47.5 \\
SegFormer-B0 & HYPER3 & LT & 60.4 & 58.4 & 86.6 & 22.6 & 43.0 & 49.9 & 53.5 \\

\bottomrule
\end{tabular}

\label{tab:ablation_labeltransfer}
\end{table*}

\begin{table*}[!tb]

\caption{Object segmentation evaluation on SpectralWaste dataset with different architectures (MiniNet-v2, SegFormer and CMX) and different modalities (RGB, HYPER, HYPER3, RGB-HYPER and RGB-HYPER3). 
}

\setlength{\tabcolsep}{5pt}
\small
\centering
\footnotesize
\begin{tabular}{lll rrrrrr c ccc}
\toprule
& & & \multicolumn{6}{c}{\textbf{IoU (\%)} \up} \\
\cmidrule(r){4-9}
\textbf{Backbone} & \textbf{Modality} & \textbf{Fusion} &\textit{Film} & \textit{Basket} & \textit{Card.} & \textit{Tape} & \textit{Filam.} & \textit{Bag} & \textbf{mIoU (\%) \up}
& \textbf{Img./s \up} & \textbf{Param. (M) \down} & \textbf{GFLOPs \down}
\\
\midrule

MiniNet-v2 & RGB & - & 63.1 & 58.9 & 55.4 & 30.6 & 10.0 & 49.2 & 44.5 &
126.7 & 0.522 & 1.343
\\
MiniNet-v2 & HYPER & - & 61.2 & 61.0 & 78.8 & 28.8 & 30.5 & 56.3 & 52.8 &
125.8 & 0.585 & 3.429
\\
MiniNet-v2 & HYPER3 & - & 58.8 & 61.9 & 69.4 & 30.2 & 23.0 & 50.5 & 49.0 &
125.9 & 0.522 & 1.431
\\
MiniNet-v2 & RGB-HYPER & early & 67.3 & 59.1 & 82.6 & 24.1 & 6.1 & 55.4 & 49.1 &
124.6 & 0.586 & 3.457
\\
MiniNet-v2 & RGB-HYPER3 & early & 57.9 & 53.3 & 69.6 & 13.5 & 6.5 & 49.6 & 41.7 &
125.0 & 0.523 & 1.459
\\
\midrule
SegFormer-B0 & RGB & - & 66.9 & 71.3 & 48.9 & 33.6 & 15.2 & 54.6 & 48.4 &
156.2 & 3.716 & 3.508
\\
SegFormer-B0 & HYPER & - & 65.4 & 63.2 & 85.2 & 21.9 & 33.1 & 57.2 & 54.3 &
152.2 & 4.062 & 6.347
\\
SegFormer-B0 & HYPER3 & - & 60.4 & 58.4 & 86.6 & 22.6 & 43.0 & 49.9 & 53.5 &
152.9 & 3.717 & 3.596
\\
SegFormer-B0 & RGB-HYPER & early & 71.3 & 62.9 & 87.5 & 21.2 & 22.0 & 56.9 & 53.6 &
155.9 & 4.067 & 6.385
\\
SegFormer-B0 & RGB-HYPER3 & early & 57.7 & 59.2 & 80.9 & 10.2 & 34.4 & 48.6 & 48.5 &
157.1 & 3.721 & 3.634
\\
\midrule
CMX-B0 & RGB-HYPER & hybrid & 77.7 & 74.9 & 80.2 & 31.1 & 20.7 & 64.5 & 58.2 &
54.7 & 11.539 & 8.365
\\
CMX-B0 & RGB-HYPER3 & hybrid & 71.7 & 71.6 & 71.7 & 27.8 & 37.7 & 59.4 & 56.6 &
55.1 & 11.193 & 5.615
\\

\bottomrule
\end{tabular}

\label{tab:experiments}
\end{table*}

\subsection{Label Transfer Evaluation}  
\label{sec:lt_eval}
To obtain a reliable evaluation of the proposed algorithm for transferring labels,  we manually annotate 20 hyperspectral images with 81 object instances, ensuring that all the classes appear in the set. 

Table~\ref{tab:manual_vs_transfer} presents the results of the proposed label transfer approach (LT) in comparison to the base manual alignment (MA) performed in the general preprocessing step. The manual alignment process involves cropping the excess image captured by each camera to align the space shown and resizing them to the same shape. The evaluation of both methods is based on the intersection over union (IoU) of the resulting masks with the set of 81 instances manually labeled in the hyperspectral images.
In the case of big objects, i.e., film, basket, cardboard, and trash bag, label transfer demonstrates an improvement on every class ranging between 9.7 and 17.2\% compared to the manual alignment, mainly due to the high volume of the segmented objects.
On the more complex classes with thin objects, i.e., video tape and filaments, there is an increase to 29.7 and 56.1\%, respectively.
In fact, in these thin objects, the baseline does not even reach 28\% of mIoU with the ground truth.
In summary, the proposed label transfer improves the average mIoU by 23.3\%  with respect to the baseline and achieves a mIoU of~79.6\%.

\begin{figure}[!tb]
    \centering
    \includegraphics{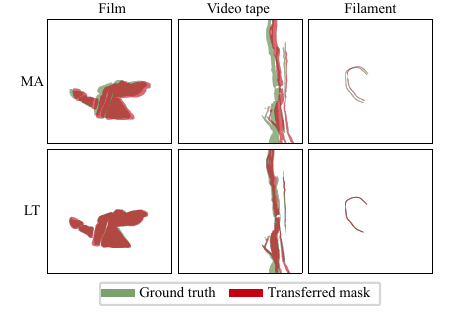}
    \caption{Qualitative results of the label transfer evaluation. The manually annotated masks are shown in green and the resulting mask from each method is in red. First row visualizes the proposed label transfer (LT) and second row the manual alignment (MA).}
    \label{fig:qualitative_resultsLT}
\end{figure}

Figure~\ref{fig:qualitative_resultsLT} shows qualitative results from both algorithms for the challenging thin classes (video tape and filaments) and the larger class in the dataset (film). The ground truth is shown in green and the resulting masks in red. The first row corresponds to MA and the second row to LT alignment. The visualization shows how the mask transferred with the proposed label transfer algorithm matches significantly better the ground truth than the manual alignment process.  

In addition, we also evaluate the impact of training the unimodal models, MiniNet-v2 and SegFormer, with the annotations resulting from each of the alternative methods, MA and LT, as labels. Table~\ref{tab:ablation_labeltransfer} summarizes this study. The results demonstrate a higher mIoU using LT than MA in every configuration analyzed. Thus, we validate that our label transfer approach generates masks that better fit the objects in the target images, hyperspectral in our case, by reducing the noise of the segmentation annotations.    

\subsection{Segmentation Architectures Evaluation}
\label{sec:evaluation}

\paragraph*{Unimodal segmentation analysis}
First, we analyze the potential of using hyperspectral data for the object segmentation task by evaluating MiniNet-v2 and SegFormer with hyperspectral or RGB data.
The results, shown in Table~\ref{tab:experiments}, demonstrate the value of using HSI information (HYPER) for object segmentation, achieving a higher mean intersection over union (mIoU) than conventional RGB images in both~architectures.

Another relevant aspect to note arises when comparing the results using all hyperspectral bands directly as input (HYPER) with the PCA reduction to three channels (HYPER3). The good results achieved with the reduced input confirm the high variance covered with just three components in the dimensionality reduction performed with PCA.

\begin{figure*}[!tb]
    \centering
    \includegraphics{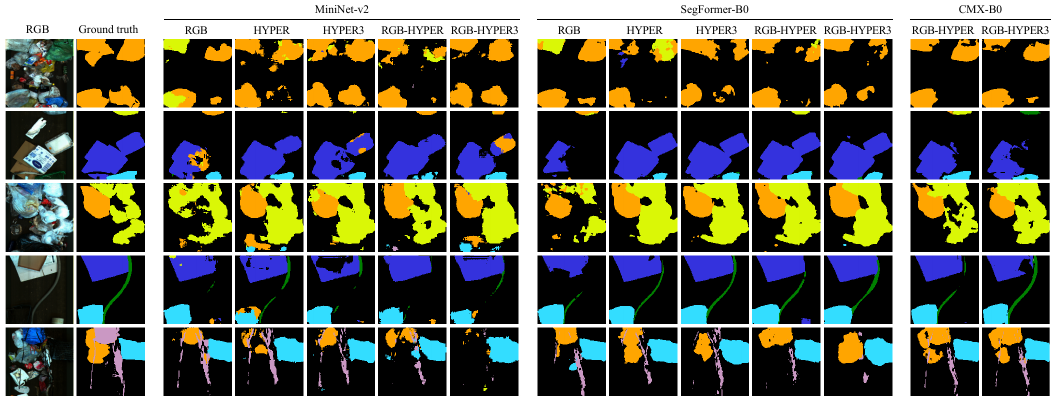}
    \includegraphics{images/legend.pdf}
    \caption{Qualitative results of the implemented architectures using the RGB, HYPER and RGB-HYPER modalities.}
    \label{fig:mosaicdata2}
\end{figure*}

Figure~\ref{fig:mosaicdata2} shows some resulting masks for a qualitative comparison between the different architectures studied. For instance, note how the filament (fourth row and green mask) is barely identified by SegFormer using RGB data. However, the resulting mask using hyperspectral information is highly accurate for this challenging class. 

\paragraph*{Multimodal segmentation analysis} Regarding the assessment of the multimodal results summarized in Table~\ref{tab:experiments}, the CMX architecture with hybrid fusion outperforms the early-fusion baselines. Considering that the images from both modalities are not perfectly aligned, it sounds reasonable that hybrid feature fusion is able to smooth out the noise and leverage the combination of information better than the early-fusion methods.

\paragraph*{Efficiency analysis}
This evaluation is conducted on a computer equipped with an AMD Ryzen 9 5950X CPU and a NVIDIA GeForce RTX 4090 GPU. 
Table~\ref{tab:experiments} examines the computational load (GFLOPs) of each configuration when processing one image at a time.
Measurements also include memory (number of parameters) and inference throughput (images/s).
Note that batch inference can improve the processing time per image.
When HYPER3 is used, we account additionally for the dimensionality reduction overhead.
The throughput results showcase the real-time running capability of the implemented architectures while a significant increase in GFLOPs occurs when using the 224-channel hyperspectral images (HYPER). Employing HYPER data as input brings slight improvements in accuracy with respect to the reduced version HYPER3 in the analyzed configurations. However, the efficiency analysis clearly suggests that the reduced version of hyperspectral imaging is a better choice, offering the best trade-off between accuracy and computational load.
Each of the architectures evaluated presents distinct trade-offs in performance and efficiency. 
MiniNet-v2 is designed to run efficiently on CPU, prioritizing low computational and memory demands (GFLOPs and number of parameters respectively). This design, optimized for CPU execution, results in slower processing speeds (images/s) on GPUs compared to SegFormer. Conversely, CMX requires significantly higher computational demands but surpasses both MiniNet-v2 and SegFormer in terms of mIoU, achieving the highest segmentation accuracy.

\section{Conclusions}
This paper has introduced SpectralWaste, the first multimodal dataset collected from a real waste sorting facility comprising RGB and HSI images. The presented dataset addresses the identification of critical objects that impact the efficiency of the sorting process. 
In the context of waste object segmentation, we have also proposed a pipeline that pays special attention to employing efficient architectures and exploiting the synergies between multiple sensing modalities.
Our comprehensive evaluation with SpectralWaste demonstrates the benefits of using RGB and HSI for waste object segmentation.
On the other hand, the low performance in segmenting some of the classes with current state-of-the-art architectures, remarks the open challenges and opportunities that SpectralWaste poses for future research in waste segmentation.

{
\balance
\bibliographystyle{IEEEtran}
\bibliography{separabib}
}

\end{document}